\documentclass{article}

\usepackage{neurips_2026}

\usepackage[utf8]{inputenc}
\usepackage[T1]{fontenc}
\usepackage{natbib}
\usepackage{hyperref}
\usepackage{url}
\usepackage{booktabs}
\usepackage{amsfonts}
\usepackage{amssymb}
\usepackage{amsmath}
\usepackage{nicefrac}
\usepackage{microtype}
\usepackage{xcolor}
\usepackage{graphicx}
\usepackage{enumitem}
\usepackage{subcaption} 

\title{SCI-CLIP: Segment-Centric Inference with Reference Memory for Training-Free Open-Vocabulary Segmentation}

\begin{document}

\maketitle

\begin{abstract}
Training-free open-vocabulary segmentation remains limited by a missing inference abstraction. Frozen vision-language features are produced at patch level, yet dense prediction requires a unit that simultaneously governs feature interaction, spatial support, contextual recovery, and retrieval-based correction. We present \textbf{SCI-CLIP}, a segment-centric inference framework built around the principle that the same region abstraction should organize all stages of dense open-vocabulary prediction. SCI-CLIP first induces a region-consistent interaction graph over frozen visual tokens, then reconstructs dense features by propagating values over this graph, augmenting them with selective cross-window support only where local evidence is insufficient. The same segment abstraction is subsequently used to construct and query an offline reference memory, aligning exemplar retrieval with the units on which prediction is made. SCI-CLIP turns frozen CLIP-style features into spatially coherent, context-aware, and retrieval-compatible dense predictions without any training. SCI-CLIP consistently improves the structural quality of dense predictions, the robustness of contextual reasoning, and the alignment of exemplar-based correction, yielding stronger open-vocabulary segmentation across eight benchmarks. Project code is available at:
\href{https://github.com/mzamini92/SCICLIP}{\texttt{github.com/mzamini92/SCICLIP}}.

\end{abstract}

\section{Introduction}

Open-vocabulary semantic segmentation (OVSS) aims to assign semantic labels to pixels from an arbitrary vocabulary, enabling dense recognition beyond the categories seen during training. A common approach is to repurpose frozen vision-language models for dense prediction: text prompts provide class embeddings, image patches provide visual features, and pixel-level logits are obtained through visual-text similarity. While attractive for its simplicity and scalability, this paradigm exposes a fundamental mismatch between image-level vision-language pretraining and region-level dense recognition. Patch tokens are optimized for global semantic alignment rather than spatially consistent object delineation; their interactions are not constrained by object extent; and ambiguous regions often require contextual or exemplar-based evidence that is not naturally available from local similarity.

These limitations point to a deeper issue: training-free OVSS is missing the right unit of inference. Dense prediction is not only a classification problem over pixels, but also a problem of \emph{where} information is allowed to propagate, \emph{how} evidence should be aggregated, \emph{when} context should be recovered beyond a local crop, and \emph{which} uncertain regions should be corrected using external exemplars. If these decisions are made at incompatible granularities, dense predictions, contextual support, and retrieval-based correction become only loosely coupled.

We introduce SCI-CLIP, a training-free framework for open-vocabulary segmentation built around a single principle: the missing abstraction is the \emph{segment} itself. SCI-CLIP treats segments not as a final post-processing artifact, but as the native support of inference. Region masks define the units over which token interactions are admissible, dense visual evidence is reconstructed, non-local support is selectively recovered, and exemplar correction is performed. This segment-centric design preserves the open-vocabulary semantics of frozen vision-language models while turning dense prediction into a spatially structured inference problem.

\begin{figure*}
    \centering
    \includegraphics[width=1.0\linewidth]{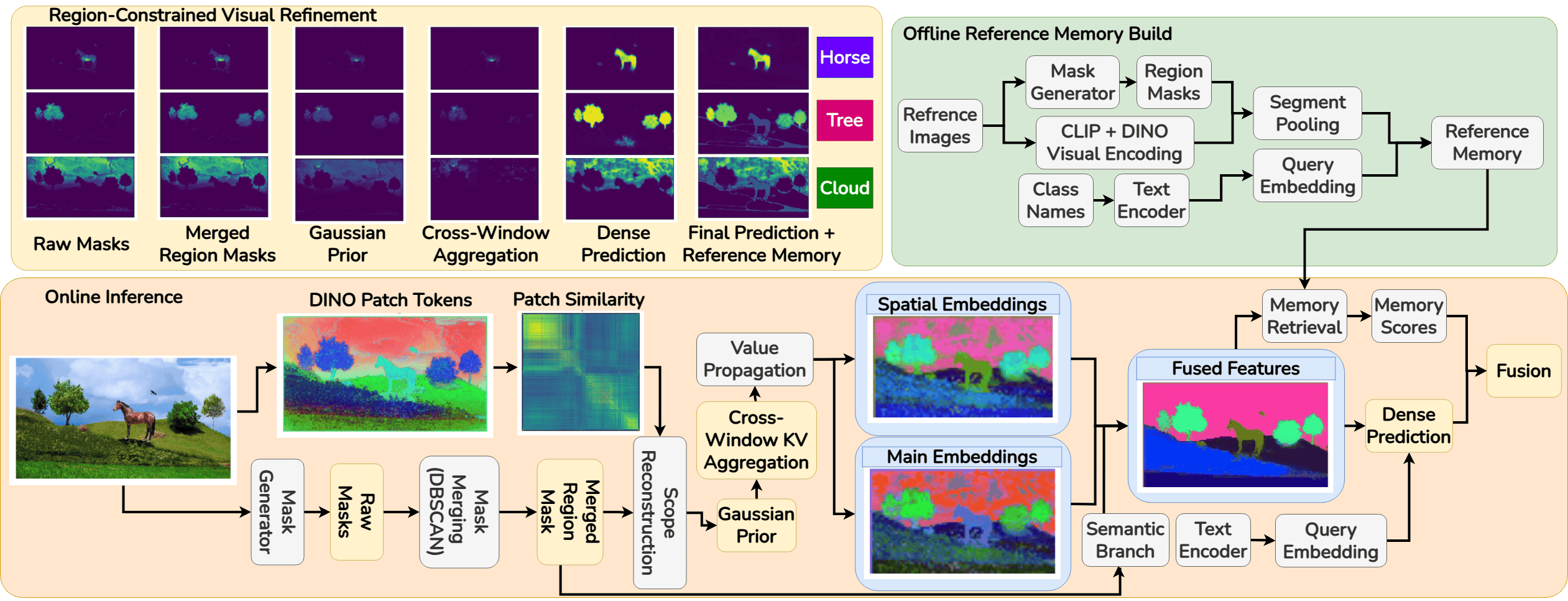}
    \caption{Overview of SCI-CLIP. SCI-CLIP consists of an online inference pathway coupled with an offline reference-memory construction pathway. During online inference, the input image is first decomposed into region masks, which are merged and used to constrain patch-to-patch interaction through scope reconstruction. A Gaussian spatial prior and cross-window KV aggregation then refine value propagation, producing main and spatial embeddings that are fused with a semantic branch to form the final dense feature map. This reconstructed feature map is matched with text-side query embeddings to produce dense predictions, while a parallel memory-retrieval branch queries an offline reference memory built from masked segment pooling over reference images and class-name embeddings. The final segmentation map is obtained by fusing dense query predictions with retrieved reference-memory scores. The top-left qualitative strip visualizes the progressive effect of raw masks, merged masks, Gaussian bias, cross-window aggregation, dense prediction, and reference-memory fusion on representative query responses.}
    \label{fig:framework}
\end{figure*}

SCI-CLIP first derives a region-level structure from the input image and uses it to define a constrained interaction graph over frozen visual tokens. It then reconstructs dense representations through region-aware value transport, spatially biased aggregation, and selective cross-window support, allowing local predictions to recover missing evidence without forfeiting segment consistency. Finally, SCI-CLIP performs reference-memory fusion at the same segment granularity, so that offline exemplar evidence and online prediction are aligned by construction rather than only at the level of final scores. Importantly, all of this is achieved without updating the vision-language backbone, making SCI-CLIP a true inference-time reformulation of training-free OVSS. Figure~\ref{fig:framework} provides an overview.

By centering all inference stages on the same segment-level support, SCI-CLIP eliminates the granularity mismatch between feature propagation, contextual reasoning, and memory-based correction that limits prior training-free approaches. The resulting system is not a frozen backbone plus several local fixes, but a coherent inference framework in which region structure determines how evidence is routed, refined, and corrected. Across standard OVSS benchmarks, this abstraction yields spatially coherent and semantically stronger predictions while retaining the flexibility of training-free vision-language inference.
\section{Related Work}
\label{sec:related}


Training-free open-vocabulary semantic segmentation has recently gained attention as an alternative to full-model adaptation, demonstrating that frozen vision--language models can be repurposed for dense prediction by reorganizing their feature representations \cite{zhang2025corrclip,lee2026looking,xuan2025reme}. Despite encouraging progress, most existing approaches still treat dense refinement, contextual recovery, and external correction as separate interventions around the same frozen predictor, rather than deriving them from a common inference abstraction.

A first line of work investigates the structure of CLIP features themselves. These methods observe that representations learned for global image--text alignment are not directly suitable for spatially coherent prediction, and therefore intervene on attention or feature interactions to improve local organization. For example, NACLIP \cite{hajimiri2025pay} redistributes attention toward local neighborhoods to reduce diffuse activations, SCLIP \cite{wang2024sclip} reformulates attention as self-correlation to enhance feature sharpness, and ClearCLIP \cite{lan2024clearclip} removes residual connections to disentangle semantic and positional information. These approaches sharpen dense features, but they do not define the region-level support on which downstream contextual recovery and retrieval should operate.

A second line of work addresses the loss of context introduced by sliding-window inference, which is commonly required for high-resolution inputs. Processing images in independent crops can lead to boundary artifacts and inconsistent predictions unless information is shared across windows \cite{lee2026looking} \cite{kim2025distilling} \cite{lan2024proxyclip}. These methods improve cross-window consistency, but the mechanism that restores context is typically not tied to the same units that govern local feature reconstruction.

A third direction incorporates external reference information at inference time. These methods augment per-image evidence by retrieving exemplars or constructing reference sets to guide dense prediction. For instance, FreeDA \cite{barsellotti2024training} and related diffusion-based approaches \cite{karazija2024diffusion} leverage text-corresponding attention masks to build reference regions aligned with semantic queries. ReME \cite{xuan2025reme} retrieves segment-level memory features and further relies on vision--language models to generate auxiliary signals, introducing additional computational overhead and sensitivity to the underlying VLM quality. While such strategies improve performance by injecting external context, they still leave open the question of whether the retrieval units are intrinsically aligned with the dense predictor that consumes them.

Our experiments demonstrate that accurate training-free segmentation depends on three tightly related factors: the quality of local feature organization, the ability to recover context across spatial partitions, and the integration of external information. However, existing methods typically treat these components independently. In contrast, SCI-CLIP unifies them within a shared segment-level representation, where region structure serves as the basis for feature refinement, context propagation, and memory integration during inference.

\section{Method}
\label{sec:method}

\subsection{Problem setup}
Let an image be denoted by $I \in \mathbb{R}^{H \times W \times 3}$ and let $\mathcal{Y} = \{y_1, \dots, y_K\}$ be an open-vocabulary label set described through text prompts. A text encoder maps prompts to normalized query embeddings
\begin{equation}
    q_k \in \mathbb{R}^{C}, \qquad k = 1, \dots, K.
\end{equation}
The goal is to predict a dense semantic map over $I$ by matching each image location to the vocabulary $\mathcal{Y}$, while allowing labels that were not fixed during pretraining of the visual encoder.

SCI-CLIP starts from a frozen CLIP-style model augmented with DINO features, then introduces three additional stages: (i) region-constrained affinity reconstruction, (ii) value reconstruction with cross-window KV aggregation, and (iii) reference-memory fusion. As shown in Figure~\ref{fig:framework}, the same region structure that constrains visual refinement online is also used to pool segment embeddings for the offline reference memory. The remaining components follow from this choice: they specify how information should be routed, reconstructed, and corrected once the support is fixed.

\subsection{Visual tokenization and dense query prediction}
The image is passed through a CLIP visual encoder with DINO patch descriptors. Let
\begin{equation}
    Z = \{z_i\}_{i=1}^{N}, \qquad z_i \in \mathbb{R}^{C},
\end{equation}
denote the patch-level visual tokens at spatial resolution $H' \times W'$ with $N = H'W'$. These tokens are later projected to dense feature maps and matched against text queries. The dense query logits are
\begin{equation}
    \ell_{i,k}^{\text{dense}} = z_i^\top q_k,
\end{equation}
which are rasterized to the image plane and normalized into dense query probabilities. This branch provides the base open-vocabulary prediction, but without additional structure it is susceptible to leakage across object boundaries and to confusion among visually similar classes.

\subsection{Region-constrained affinity reconstruction}
\label{sec:scope}
SCI-CLIP uses an auxiliary mask generator to build a region decomposition of the image. Let $m_i$ denote the region identifier of patch $i$ after mapping masks to the patch grid.

We first compute a DINO-based patch similarity matrix
\begin{equation}
    S_{ij} = d_i^\top d_j,
\end{equation}
where $d_i$ is the normalized DINO descriptor at patch $i$. This raw affinity is then \emph{scope-constrained} using the region structure. Patches are allowed to interact if they belong to the same region, while unsegmented regions retain a weaker data-dependent allowance based on above-average similarity. Formally, we define a binary admissibility mask $A \in \{0,1\}^{N \times N}$ and set
\begin{equation}
    \widetilde{S}_{ij} =
    \begin{cases}
        S_{ij}, & A_{ij} = 1,\\
        -\infty, & A_{ij} = 0.
    \end{cases}
\end{equation}
This converts unconstrained global affinity into region-aware interaction, which sharply reduces cross-object mixing.

\paragraph{Mask merging.}
Region proposals may be overly fragmented. SCI-CLIP merges masks whose mean region features are highly similar. Specifically, for each region we average DINO descriptors within that region, normalize the resulting vectors, and cluster them with DBSCAN on a precomputed distance matrix. Regions assigned to the same cluster are merged before the scope mask is constructed. This step reduces redundant region boundaries and produces a cleaner interaction graph.

\paragraph{Gaussian spatial prior.}
Region constraints alone do not encode locality. SCI-CLIP adds a Gaussian spatial prior over patch coordinates:
\begin{equation}
    G_{ij} = \exp\left(-\frac{\|p_i - p_j\|_2^2}{2\sigma^2}\right),
\end{equation}
where $p_i$ is the spatial coordinate of patch $i$. After normalizing $G$, we obtain the final refined affinity
\begin{equation}
    \widehat{S} = \widetilde{S} + \lambda_g G,
\end{equation}
applied only on valid entries. The corresponding attention weights are
\begin{equation}
    W = \mathrm{softmax}(\tau \widehat{S}),
\end{equation}
where $\tau$ is a temperature parameter. The Gaussian prior acts as a local regularizer: it preserves the region-constrained graph while slightly favoring nearby patches, which is particularly useful for contiguous stuff regions and spatially coherent objects.

\subsection{Value reconstruction}
\label{sec:value}
SCI-CLIP next reconstructs value features using the refined attention matrix $W$. In the last visual transformer stage, it obtains projected value tokens from the CLIP attention block, separates them into a main branch and a spatial branch, and propagates them using the refined affinity:
\begin{equation}
    V^{\text{main}}_{\text{loc}} = W V^{\text{main}}, \qquad
    V^{\text{spatial}}_{\text{loc}} = W V^{\text{spatial}}.
\end{equation}
This process converts pairwise affinity into region-aware feature transport. The main branch preserves the original semantic representation, while the spatial branch aggregates lower-layer spatial evidence collected from multiple transformer blocks.

\paragraph{Cross-window KV aggregation.}
Sliding-window inference is often necessary for high-resolution images, but it fragments context: once the affinity graph is constrained within a crop, feature propagation can become excessively local and fail to recover evidence that lies just outside the current window. SCI-CLIP addresses this with \emph{cross-window KV aggregation}. After computing local reconstructed values inside each crop, it assembles a global bank by projecting overlapping windows back onto the full feature lattice and averaging their aligned key-value representations. For a token in the current crop, SCI-CLIP then queries this bank with its DINO descriptor, selects the most compatible cross-window entries, and forms a retrieved value vector through temperature-scaled softmax weighting. Concretely, for token $i$ with local descriptor $d_i$, global keys $\{k_j\}$, and global values $\{v_j\}$, the retrieved branch is
\begin{equation}
    a_{ij}^{\text{kv}} = \frac{\exp\left(\tau_{\text{kv}} d_i^\top k_j\right)}{\sum_{j'} \exp\left(\tau_{\text{kv}} d_i^\top k_{j'}\right)},
    \qquad
    V^{\text{main}}_{\text{glob}, i} = \sum_j a_{ij}^{\text{kv}} v_j,
\end{equation}
and analogously for the spatial branch. The retrieved global output is then blended with the local reconstruction:
\begin{equation}
    V^{\text{main}}_{\text{blend}} = (1-\lambda_{\text{kv}}) V^{\text{main}}_{\text{loc}} + \lambda_{\text{kv}} V^{\text{main}}_{\text{glob}},
\end{equation}
and analogously for the spatial branch. This design is important for SCI-CLIP: the model first suppresses harmful global mixing through scope reconstruction, then selectively reintroduces non-local evidence only through retrieved value transport. In other words, SCI-CLIP does not revert to unconstrained global attention; it adds a structured long-range correction path that respects the local reconstruction stage while recovering context across crop boundaries.

\paragraph{Semantic branch.}
In parallel, SCI-CLIP constructs a semantic branch from mask-conditioned tokens. Region masks are expanded to the feature grid and used to aggregate a bank of mask-conditioned token representations. These semantic features provide coarse region-level evidence complementary to the main and spatial value branches. After layer normalization and projection, the three branches are combined as: 
\begin{equation}
    z_i = z_i^{\text{main}} + \alpha z_i^{\text{spatial}} + \beta z_i^{\text{semantic}},
\label{eq:branch_fusion}
\end{equation}
where $\alpha$ and $\beta$ control the relative contributions of the auxiliary branches. The resulting feature map is then matched to query embeddings to form dense open-vocabulary predictions. In practice, this reconstructed representation is visibly more coherent than raw token features and aligns better with semantic extent.

\subsection{Reference-memory construction}
\label{sec:memory}
Dense prediction alone is not always reliable, especially when several classes share similar appearance. To address this, SCI-CLIP constructs an offline \emph{reference memory} of region-level exemplars. For each reference image, the same mask generator produces instance or region masks. SCI-CLIP extracts the reconstructed visual features, pools them over each region, and stores the resulting segment embeddings:
\begin{equation}
    e_r = \frac{1}{|r|} \sum_{i \in r} z_i, \qquad e_r \leftarrow \frac{e_r}{\|e_r\|_2}.
\end{equation}
Each reference segment is associated with the query labels present in the underlying image, producing a segment-to-label incidence matrix. The memory additionally stores text-side label features and optional class prototypes derived from high-confidence segments.

The key design choice is that the same segmentation abstraction is used both when constructing reference segments and when computing retrieval at test time. This alignment makes the memory semantically meaningful: retrieved exemplars correspond to the same kind of region units that appear during inference.

\subsection{Reference-memory retrieval and fusion}
At test time, SCI-CLIP extracts segment embeddings from the current image using the same region masks and computes visual affinity to the stored reference memory:
\begin{equation}
    A^{\text{seg}} = \mathrm{softmax}\left(\gamma_s E E_{\text{mem}}^\top\right),
\end{equation}
where $E$ contains test-image segment embeddings, $E_{\text{mem}}$ contains memory embeddings, and $\gamma_s$ is a retrieval scale. In parallel, label-side affinity between memory label features and active query embeddings is computed as:
\begin{equation}
    A^{\text{label}} = \mathrm{softmax}\left(\gamma_\ell Q_{\text{mem}} Q^\top\right).
\end{equation}
These are composed to obtain query scores for each test segment:
\begin{equation}
    P^{\text{mem}}_{\text{seg}} = A^{\text{seg}} Y_{\text{mem}} A^{\text{label}},
\end{equation}
where $Y_{\text{mem}}$ is the segment-to-label incidence matrix. Segment-level memory scores are then rasterized back to pixels according to the test-image region masks, producing a dense memory-induced query map $P^{\text{mem}}$.
Finally, SCI-CLIP fuses dense and memory predictions:
\begin{equation}
    P^{\text{final}} =
    \frac{\lambda_d P^{\text{dense}} + \lambda_m P^{\text{mem}}}{\lambda_d + \lambda_m},
\end{equation}
where $\lambda_d$ and $\lambda_m$ are the dense and memory weights. This late fusion is deliberately simple: the dense branch preserves open-vocabulary flexibility, while the reference memory supplies exemplar-based correction for uncertain regions.


\section{Experiments}
\label{sec:results}
\paragraph{Benchmark Datasets.} We perform a comprehensive evaluation across eight widely used segmentation benchmarks, following standard protocols. These datasets are grouped based on whether a background class is included. (i) With background class: This group includes PASCAL VOC (V21) \cite{everingham2011pascal}, PASCAL Context (C60) \cite{mottaghi2014role}, and COCO Object (Object) \cite{caesar2018coco}, where background regions are explicitly modeled as a separate class. (ii) Without background class: This category consists of PASCAL VOC20 (V20) \cite{everingham2011pascal}, PASCAL Context59 (C59) \cite{mottaghi2014role}, COCO Stuff (Stuff) \cite{caesar2018coco}, Cityscapes (City) \cite{cordts2016cityscapes}, and ADE20K (ADE) \cite{zhou2019semantic}, where evaluation is performed only over foreground semantic categories. Analysis of the retrieval hyperparameters is provided in \ref{sec:retrieval_sensitivity}.

\begin{table*}[htbp]
\centering
\small
\caption{Hyperparameters (left) and branch-weight sensitivity (right).}

\begin{subtable}{0.50\textwidth}
\centering
\setlength{\tabcolsep}{5pt}
\caption{Dataset-specific hyperparameters.}
\begin{tabular}{lccccccc}
\toprule
Parameter & ADE & City & Obj/Stf & C59/60 & V20 & V21 \\
\midrule
$\lambda_d$ & 0.5 & 0.9 & 0.8 &0.7 & 0.9 & 0.8 \\
$\lambda_m$ & 0.5 & 0.1 &  0.2 & 0.3 & 0.1 & 0.2 \\
Seg. scale & 60.0 & 60.0 & 60.0 & 60.0  & 60.0 & 20.0 \\
Label scale & 40.0 & 40.0 &  60.0 &  40.0 & 40.0 & 60.0 \\
\bottomrule
\end{tabular}
\end{subtable}
\hfill
\begin{subtable}{0.42\textwidth}
\centering
\setlength{\tabcolsep}{4pt}
\caption{Sensitivity to $(\alpha,\beta)$.}
\begin{tabular}{lccccc}
\toprule
$(\alpha,\beta)$ & ADE & City & V20 & V21 & Avg. \\
\midrule
$(0.0,0.5)$ & 29.1 & 42.0 & 89.4 & 74.3 & 58.7 \\
$(0.5,0.5)$ & 29.6 & 44.0 & 90.0 & 75.5 & 59.8 \\
$(1.0,0.5)$ & \textbf{29.8} & 44.7 & \textbf{90.3} & 75.9 & \textbf{60.2} \\
$(1.5,0.5)$ & 29.5 & 44.7 & 89.8 & 75.7 & 59.9 \\
$(1.0,0.0)$ & 29.3 & \textbf{45.2} & 89.3 & 74.5 & 59.6 \\
$(1.0,1.0)$ & 29.3 & 43.7 & 89.8 & \textbf{76.1} & 59.7 \\
\bottomrule
\end{tabular}
\end{subtable}

\label{tab:combined_tables}
\end{table*}

\paragraph{Implementation and Metric.} We instantiate SCI-CLIP with different CLIP vision towers, including ViT-B/16, ViT-L/14 \cite{radford2021learning} and ViT-H/14 \cite{cherti2023reproducible}. Aside from the backbone choice, the architectural components are kept fixed; the dataset-specific late-fusion and retrieval calibration values are listed in Table~\ref{tab:combined_tables}.a. $\tau$ has been set to 4.0 and the Gaussian prior weight has also been set to 0.25 on all settings. For dense feature extraction, we experiment using DINO (B/8 and S/8) \cite{caron2021emerging} and DINOv2 (B/8) \cite{oquab2023dinov2} backbone configuration. For mask generation, we adopt SAM2 \cite{ravi2024sam}, Mask2Former \cite{cheng2022masked}, EoMT \cite{kerssies2025your}, or SegFormer \cite{xie2021segformer}. We evaluate semantic segmentation performance using mean Intersection over Union (mIoU) and report all results directly from the model outputs, without applying any additional post-processing. 


\paragraph{Branch-weight sensitivity.}
\label{sec:branch_weight_sensitivity}

\begin{table*}[t]
    \centering
    \small
    \setlength{\tabcolsep}{7pt}
    \caption{Main comparison with recent training-free OVS methods. We report mIoU on eight benchmarks.}

    \begin{tabular}{lcccccccc}
        \toprule
        Method& V21 & PC60 & Object & V20 & PC59 & Stuff & City & ADE \\

        \midrule
        \multicolumn{9}{c}{\emph{CLIP B/16}} \\
        \midrule

        OVDiff \cite{karazija2024diffusion}
        & 66.3 & 29.7 & 34.6 & 80.9 & 32.9 & 20.3 & 23.4 & 14.1 \\
        DIH-CLIP \cite{duan2025dih}&64.2&36.0&37.4&84.9&39.7&26.7&40.2&19.6\\
        SFP \cite{jin2025feature} &56.8& 32.3 &32.1 &83.4& 36.0& 24.0 &34.1 &18.1 \\  
        FSA \cite{chi2025plug}  
        & 63.7&36.1&38.0&82.3&39.9&27.0&38.8&20.5 \\

        CASS \cite{kim2025distilling} & 62.3 & 34.0 & 36.0 & 87.6 & 37.6 & 25.4 & 36.2 & 18.8\\
        Trident \cite{shi2025harnessing} & 67.1 & 38.6 & 41.1 & 84.5 & 42.2 & 28.3 & 42.9 & 21.9 \\
        CorrCLIP \cite{zhang2025corrclip}
        & 72.0 & 41.1 & 43.8 & 87.8 & 46.4 & 31.1 & \textbf{45.3} & 23.7 \\  
        ResCLIP \cite{yang2025resclip} 
        & 61.1 & 33.5 & 35.0 & 86.0 & 36.8 & 24.7 & 35.9  & 18.0 \\      
        DeCLIP \cite{wang2025declip}&59.7&35.3&36.4&85.0&39.2&25.3&32.8&21.9\\
        GLA-CLIP \cite{lee2026looking}
        & 63.2 & 35.8 & 37.7 & 81.5 & 39.7 & 26.8 & 38.5 & 20.3 \\

        PEARL \cite{pei2026pearl}
        & 64.1 & 35.1 & 37.3 & 86.9 & 38.5 & 26.3 & 37.6 & 19.4 \\

        SCI-CLIP 
        & \textbf{75.9} & \textbf{46.1} & \textbf{45.2} & \textbf{90.3} & \textbf{49.6} & \textbf{33.8} & 44.7 & \textbf{29.8} \\
        \midrule
        \multicolumn{9}{c}{\emph{CLIP L/14}} \\
        \midrule
        ProxyCLIP \cite{lan2024proxyclip}&60.6&34.5&39.2&83.2&37.7&25.6&40.1&22.6\\
        Freeda \cite{barsellotti2024training}&55.4&38.3&37.4&87.9&43.5&28.8&36.7&23.2\\
        FSA \cite{chi2025plug} &67.9&36.3&40.2&85.7&40.5&27.3&43.6&24.5\\
        ResCLIP \cite{yang2025resclip}
        & 54.1 & 30.9 & 32.5 & 85.5 & 34.5 & 23.4  & 33.7 & 18.2\\
        CorrCLIP \cite{zhang2025corrclip}&71.2&41.6&45.7&90.7&46.1&30.8&46.3&26.7\\
        SCI-CLIP&\textbf{75.7}&\textbf{47.0}&\textbf{47.7}&\textbf{91.2}&\textbf{52.3}&\textbf{35.0}&\textbf{47.8}&\textbf{34.4}\\
        \midrule
        \multicolumn{9}{c}{\emph{CLIP H/14}} \\
        \midrule
        ProxyCLIP \cite{lan2024proxyclip}&65.0&35.4&38.6&83.3&39.6&26.8&42.0&24.2\\
        FSA \cite{chi2025plug} &67.9&36.3&40.2&85.7&40.5&27.3&43.6&24.5\\
        Trident \cite{shi2025harnessing} &70.8&40.1&42.2&88.7&44.3&28.6&47.6&26.7\\
        CorrCLIP \cite{zhang2025corrclip}&75.5&41.7&48.1&91.4&46.9&32.1&47.3&28.1\\
        SCI-CLIP&\textbf{76.3}&\textbf{45.4}&\textbf{49.4}&\textbf{91.5}&\textbf{49.8}&\textbf{37.1}&\textbf{49.9}&\textbf{31.9}\\
                
        \bottomrule
    \end{tabular}
    \label{tab:main_results}
\end{table*}
Table~\ref{tab:combined_tables}.b. studies the branch weights in Eq.~\eqref{eq:branch_fusion}. Suppressing the spatial branch by setting $\alpha=0$ consistently degrades performance, indicating that spatial reconstruction provides an essential complementary signal beyond the main branch alone. Increasing $\alpha$ from $0$ to $1.0$ improves the average mIoU, while further increasing it to $1.5$ yields no additional benefit. The semantic branch is also helpful, but its effect is more dataset-dependent: $\beta=0$ slightly favors City, whereas $\beta=1.0$ improves V21. We used $(\alpha,\beta)=(1.0,0.5)$ as default choice in all experiments.

\subsection{Comparison with State-of-the-Art Methods}
\label{sec:main_comparison}
Table~\ref{tab:main_results} reports the principal comparison against recent training-free open-vocabulary segmentation methods. Across the upper comparison block, SCI-CLIP achieves the best result on seven of the eight benchmarks, including strong gains on VOC21, PC60, VOC20, PC59, Stuff, and ADE. The only exception is City, where CorrCLIP remains slightly higher. The lower comparison block shows that this advantage persists under a stronger evaluation setting, where SCI-CLIP again leads every listed baseline. Taken together, the table indicates that SCI-CLIP improves not only peak object-centric performance, but also the harder context-heavy benchmarks where dense open-vocabulary inference is typically more brittle.

\begin{table*}[htbp]
\centering
\caption{Segmentation performance across various visual feature backbone and mask generators.}

\begin{subtable}{0.48\textwidth}
\centering
\caption{Dense visual feature backbone comparison}
\begin{tabular}{lccccc}
\hline
Model 
& \rotatebox{90}{V21} 
& \rotatebox{90}{V20} 
& \rotatebox{90}{C59} 
& \rotatebox{90}{C60} 
& \rotatebox{90}{City} \\
\hline
DINOv2$_B$  & 74.4 & 89.9 & 49.8 & 46.2 & 43.5  \\
DINOv1$_B$  & 75.9 & 90.3 & 49.6 & 46.1 & 44.7 \\
DINOv1$_S$  & 75.9 & 89.4 & 45.2 & 48.4 & 43.8 \\
MAE$_B$     & 69.2 & 87.7 & 41.6 & 44.9 & 37.3 \\
MAE$_L$     & 72.6 & 89.0 & 47.7 & 44.2 & 38.8 \\
\hline
\end{tabular}
\end{subtable}
\hfill
\begin{subtable}{0.46\textwidth}
\centering
\caption{Mask generator comparison. SF denotes SegFormer and M2F denotes Mask2Former.}
\begin{tabular}{lccccc}
\hline
Mask 
& \rotatebox{90}{V21} 
& \rotatebox{90}{V20} 
& \rotatebox{90}{C59} 
& \rotatebox{90}{C60} 
& \rotatebox{90}{City} \\
\hline
SF   & 68.1 & 82.5 & 49.5 & 44.2 & 42.2 \\
SAM2 & 75.9 & 90.3 & 49.6 & 46.1 & 44.7 \\
M2F  & 75.9 & 90.0 & 54.1 & 48.5  & 42.6 \\
EOMT & 76.0 & 90.5 & 49.9 &45.6 & 46.7 \\
\hline
\end{tabular}
\end{subtable}

\label{tab:combined_results}
\end{table*}
\subsection{Vision tower and mask generator comparison}
\label{sec:model_mask_comparison}

Table~\ref{tab:combined_results}.a. summarizes the performance of SCI-CLIP with different visual feature backbones, while Table~\ref{tab:combined_results}.b. shows the effect of different mask generators. The left half of the table shows that the DINO-family backbones remain the strongest overall choices, especially on VOC-style and Cityscapes benchmarks, while MAE backbones remain viable but consistently weaker in dense segmentation quality. The right half shows that SCI-CLIP is not tied to a single region source: SAM2 is a reliable default, EOMT improves City, and Mask2Former is especially strong on Context benchmarks. This pattern reinforces one of the practical strengths of SCI-CLIP: the segment-centric inference pipeline remains effective across multiple segmentation front ends, while still benefiting from stronger visual encoders when they are available.




\begin{figure*}[t]
    \centering
    \includegraphics[width=0.95\linewidth]{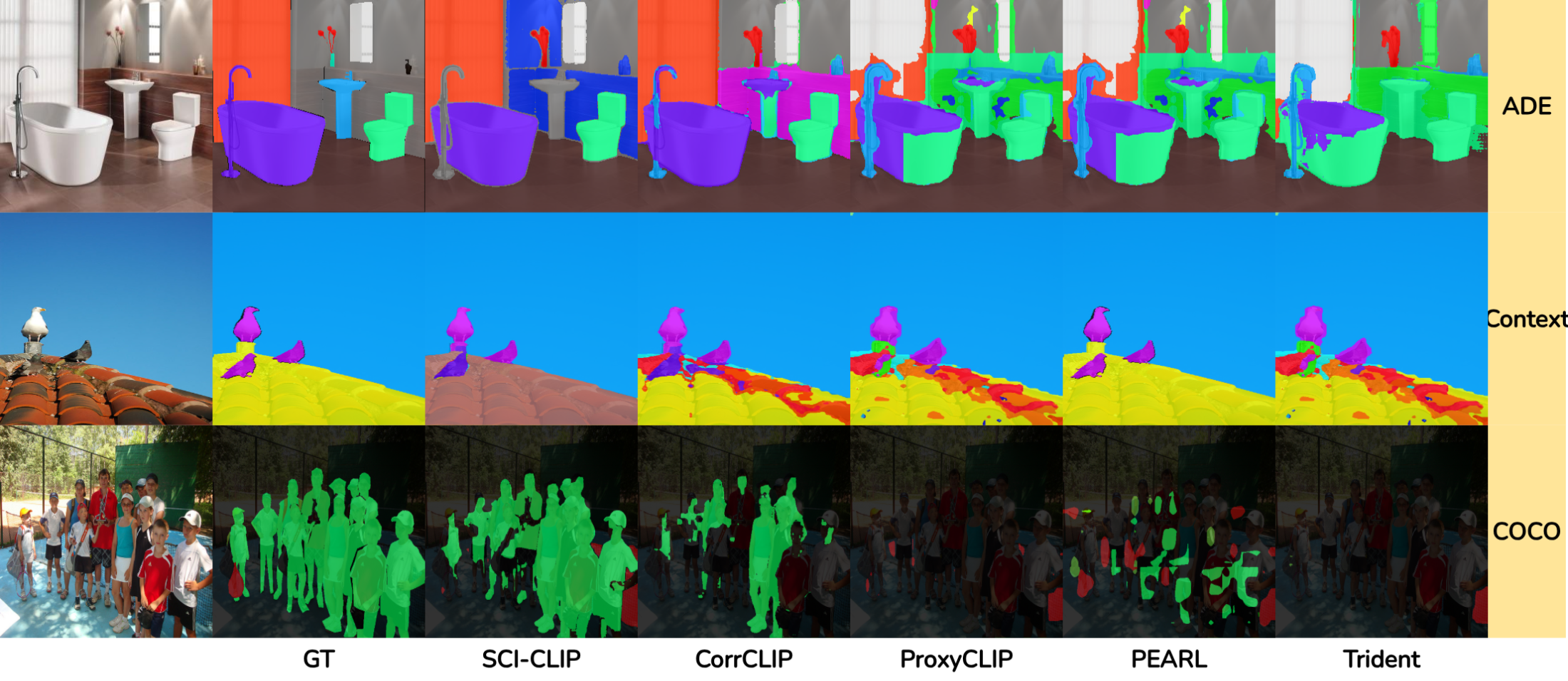}
    \caption{Qualitative comparison of SCI-CLIP with existing segmentation methods. More qualitative examples are shown in the 
    Appendix~\ref{sec:qualitatives}.
    }
    \label{fig:SCI-CLIP_qual}
\end{figure*}
\vspace{-2pt}
\subsection{Qualitative Comparison}

We present qualitative comparisons of SCI-CLIP against several state-of-the-art segmentation models across diverse benchmarks, including ADE20K, Pascal Context, and COCO. As shown in Fig.~\ref{fig:SCI-CLIP_qual}, SCI-CLIP consistently produces more spatially coherent and semantically accurate segmentation maps. In structured indoor scenes, SCI-CLIP preserves fine object boundaries (e.g., bathtub, sink, and toilet) while reducing label bleeding and over-smoothing effects observed in competing methods. In challenging scenarios with irregular object shapes, SCI-CLIP maintains better object completeness and avoids fragmentation, accurately capturing elongated structures that other models often truncate or merge with the background. Furthermore, in crowded real-world scenes, SCI-CLIP demonstrates improved robustness to occlusion and instance-level ambiguity, yielding cleaner separation of multiple objects without collapsing them into coarse regions or introducing spurious predictions. These results highlight SCI-CLIP's ability to balance global semantic consistency with local detail preservation, leading to superior visual segmentation quality across varying scene complexities.

\begin{figure*}[htbp]
    \centering
    \includegraphics[width=0.95\textwidth]{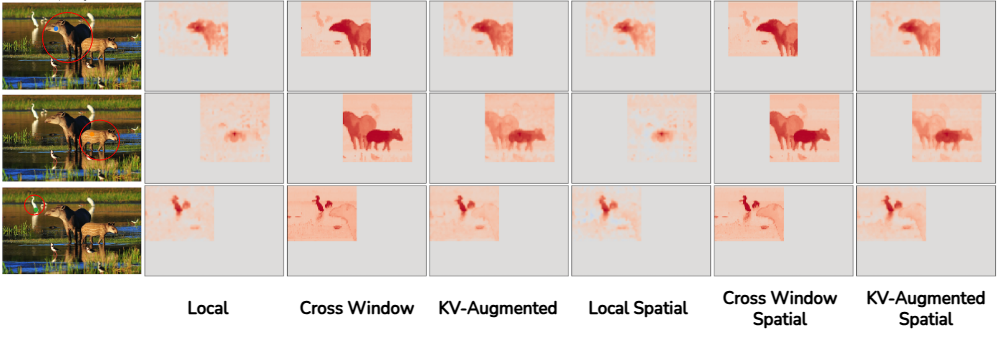}
        \caption{Cross-window KV aggregation restores context across crop boundaries. Each row corresponds to a selected query point. 
        }

    \label{fig:cross_window_kv}
\end{figure*}


Figure~\ref{fig:cross_window_kv} visualizes the effect of cross-window KV aggregation on the reconstructed main and spatial embeddings. Local reconstruction alone can remain crop-dependent, especially near window boundaries or when supporting evidence lies just outside the current crop. The retrieved cross-window branch recovers compatible long-range cues from overlapping windows, and the resulting KV-augmented embeddings strengthen semantically relevant regions without reverting to unconstrained global mixing. This qualitative behavior matches the intended role of cross-window KV aggregation in SCI-CLIP: it complements region-constrained local reconstruction with a structured long-range correction pathway.

\subsection{Stagewise ablation}
\label{sec:stagewise_ablation}
Table~\ref{tab:stagewise_ablation} provides a stagewise decomposition of SCI-CLIP across representative benchmarks. First, \emph{scope reconstruction} delivers the largest non-retrieval gain, improving the average mIoU from $45.8$ to $49.5$ and yielding substantial jumps on all datasets. This behavior is consistent with the core motivation of SCI-CLIP: unconstrained dense token interaction is a major source of error, and enforcing region-consistent affinity already resolves a large fraction of that failure mode. Additional leave-one-out ablation results for SCI-CLIP are provided in the Appendix \ref{sec:loo_ablation}.
\begin{table}[htbp]
    \centering
    \small
    \setlength{\tabcolsep}{5.5pt}
    \caption{Stagewise ablation of SCI-CLIP. We progressively enable its components following the inference pipeline and report mIoU on six representative benchmarks. SR denotes Scope Reconstruction; MM, Mask Merging; GP, Gaussian Prior; CW-KV, Cross-Window Key--Value Aggregation; and RM, Reference Memory.
    }
    \begin{tabular}{lccccccc}
        \toprule
        Stage & C60 & City & V20 & V21 & Ade20K & Stuff&Avg. \\
        \midrule
        Dense & 38.1 & 36.9 & 86.1 & 67.1 & 21.4 & 28.2& 45.8 \\
        + SR & 41.8 & 44.0 & 88.0 & 72.6 & 23.2 &30.6& 49.5 \\
        + GP & 42.0 & 44.1 & 88.3 & 72.8 &23.4& 30.7& 49.6\\
        + GP & 42.1 & 41.4 & 88.7 & 72.7 & 22.5& 30.5 & 49.2 \\
        +CW & 42.2 & 44.2 & 88.9 & 72.8 & 23.3&30.9& 49.8 \\
        + RM & \textbf{46.1} & \textbf{44.7} & \textbf{90.3} & \textbf{75.9} & \textbf{29.8}&\textbf{33.8}&\textbf{53.4} \\
        \bottomrule
    \end{tabular}
    \label{tab:stagewise_ablation}
\end{table}

\emph{Mask merging} produces a smaller but stable improvement on top of scope reconstruction, indicating that cleaning fragmented region proposals remains useful even after the affinity matrix has been region-constrained. The \emph{Gaussian prior} is more nuanced. It improves C60 and VOC20, but slightly degrades City and VOC21 when applied in isolation, suggesting that locality bias alone can become overly restrictive when long-range evidence is needed. This is precisely where \emph{cross-window KV aggregation} becomes important: once cross-window retrieval is added, the average mIoU rises to $49.8$, the best result among all non-memory stages. In other words, the Gaussian prior sharpens local support, while cross-window KV restores the missing long-range context.

The final stage, \emph{reference-memory fusion}, produces the strongest overall result, reaching $52.9$ average mIoU and improving over the dense baseline by $7.6$ points. The gain is particularly pronounced on V21 and C60, where dense open-vocabulary predictions remain more ambiguous and exemplar-based correction is most beneficial. Importantly, the ablation does not support a ``retrieval-only'' interpretation of SCI-CLIP. Instead, it shows that the method derives its strength from a layered progression: region-constrained visual refinement stabilizes the dense feature space, cross-window KV aggregation recovers context lost under crop-wise inference, and reference memory then acts as a high-precision corrective signal on top of that improved representation.

\section{Conclusion}

We presented SCI-CLIP, a segment-centric inference framework for OVSS. SCI-CLIP determines where information may propagate, how dense features are reconstructed, how non-local context is selectively recovered, and how external exemplars are stored and applied. This reformulation resolves the granularity mismatch that limits prior training-free OVSS pipelines, where feature refinement, contextual aggregation, and retrieval-based correction are often defined on incompatible units. Empirically, this yields a model that is more spatially coherent, more robust under crop-wise inference, and better aligned with exemplar-based correction, while remaining fully training-free. More broadly, SCI-CLIP suggests that progress in frozen vision-language dense prediction depends not only on stronger backbones or stronger retrieval, but on identifying the correct inference abstraction on which those operations should act.




\newpage
\bibliographystyle{plainnat}  
\bibliography{references}
\appendix

\begin{figure*}[htbp]
    \centering
    \includegraphics[width=0.95\linewidth]{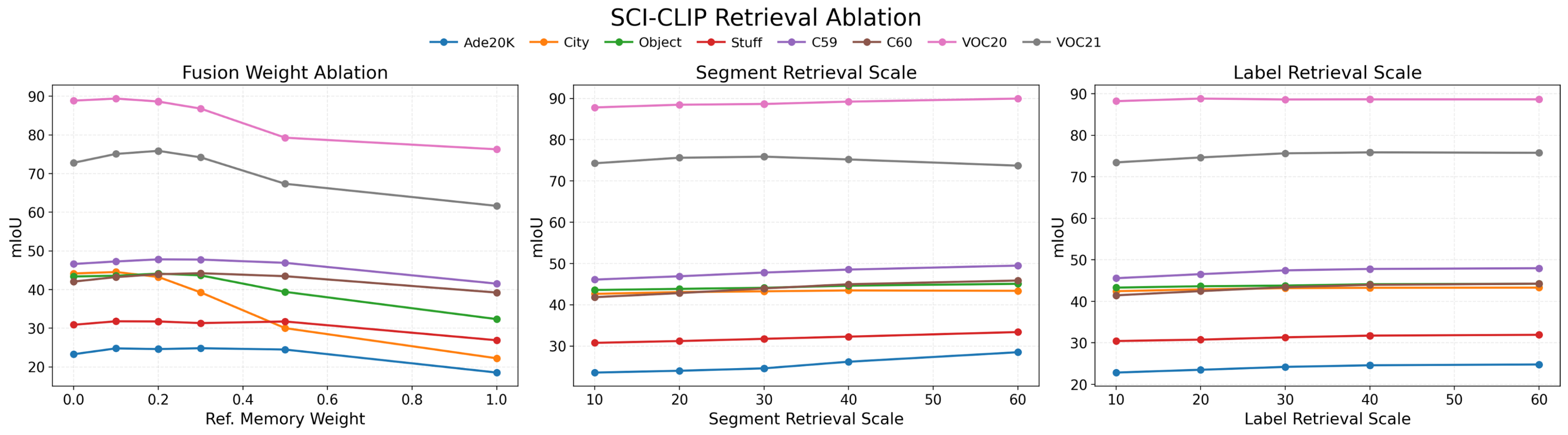}
    \caption{Sensitivity of SCI-CLIP to retrieval hyperparameters. We vary the late-fusion reference-memory weight together with the segment- and label-retrieval scales, while keeping the dense feature reconstruction pipeline fixed. The plots report mIoU across eight benchmarks. SCI-CLIP is most sensitive to the memory-fusion weight, where overly aggressive retrieval can degrade performance, whereas the segment and label retrieval scales exhibit smoother behavior over a broad range of values.}
    \label{fig:retrieval_ablation}
\end{figure*}

\section{Retrieval hyperparameter sensitivity}
\label{sec:retrieval_sensitivity}
Figure~\ref{fig:retrieval_ablation} summarizes the sensitivity of SCI-CLIP to its retrieval hyperparameters. Importantly, this study concerns the \emph{late retrieval stage}, not the internal branch weights used in dense feature reconstruction. The dominant effect comes from the fusion weight assigned to reference memory: moderate values consistently outperform both dense-only prediction and overly retrieval-heavy settings. By contrast, the segment and label retrieval scales produce smoother trends, indicating that SCI-CLIP is comparatively robust to the sharpness of the retrieval distributions once the dense-memory balance is calibrated properly. This behavior is consistent with the design of SCI-CLIP, in which reference memory acts as a corrective signal rather than a replacement for dense prediction.

\section{Qualitative comparison with Prior Methods}
\begin{figure*}[htbp]
    \centering
    \includegraphics[width=0.95\linewidth]{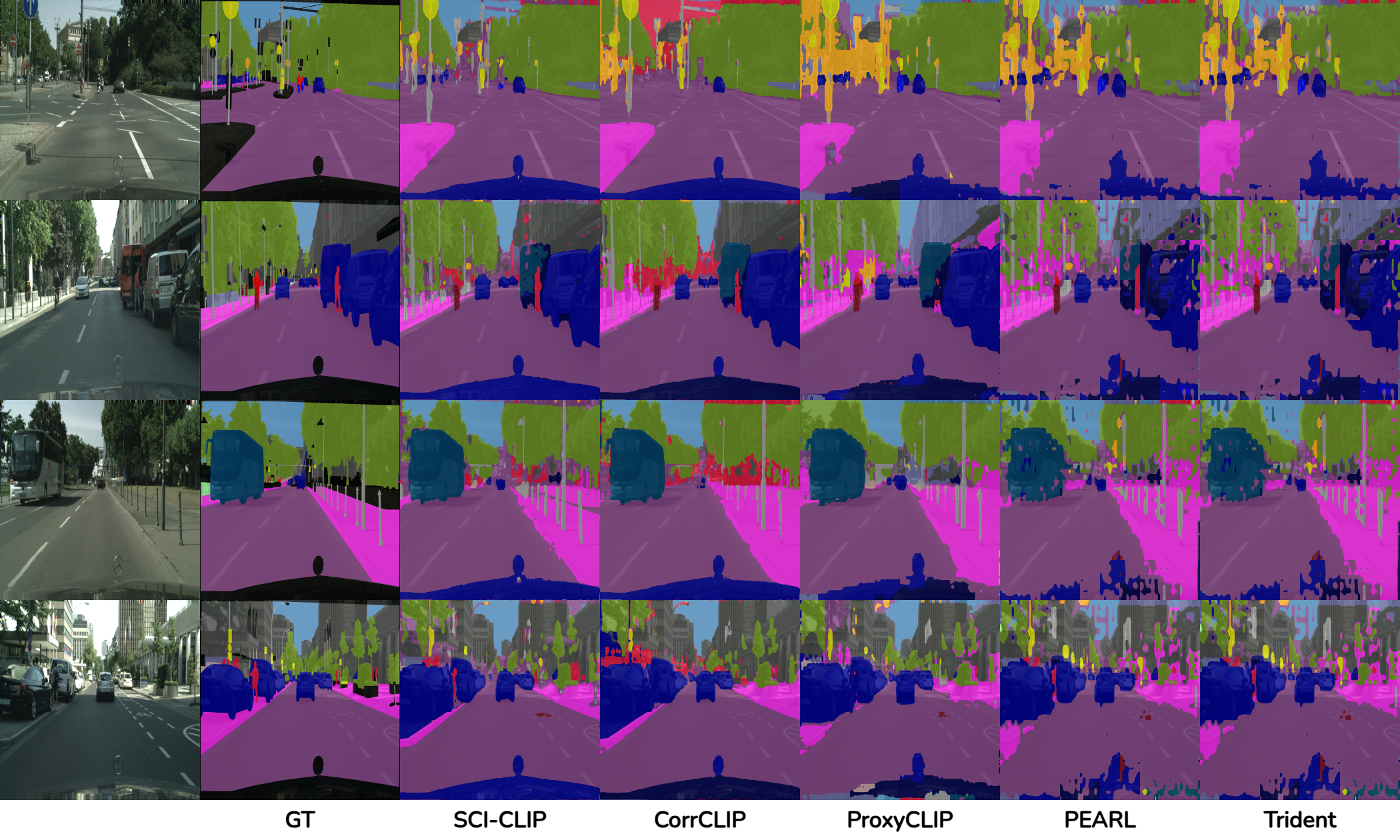}
    \caption{More qualitative comparison of our method, and the other four methods on City Scapes.}
    \label{fig:city}
\end{figure*}
\begin{figure*}[htbp]
    \centering
    \includegraphics[width=0.95\linewidth]{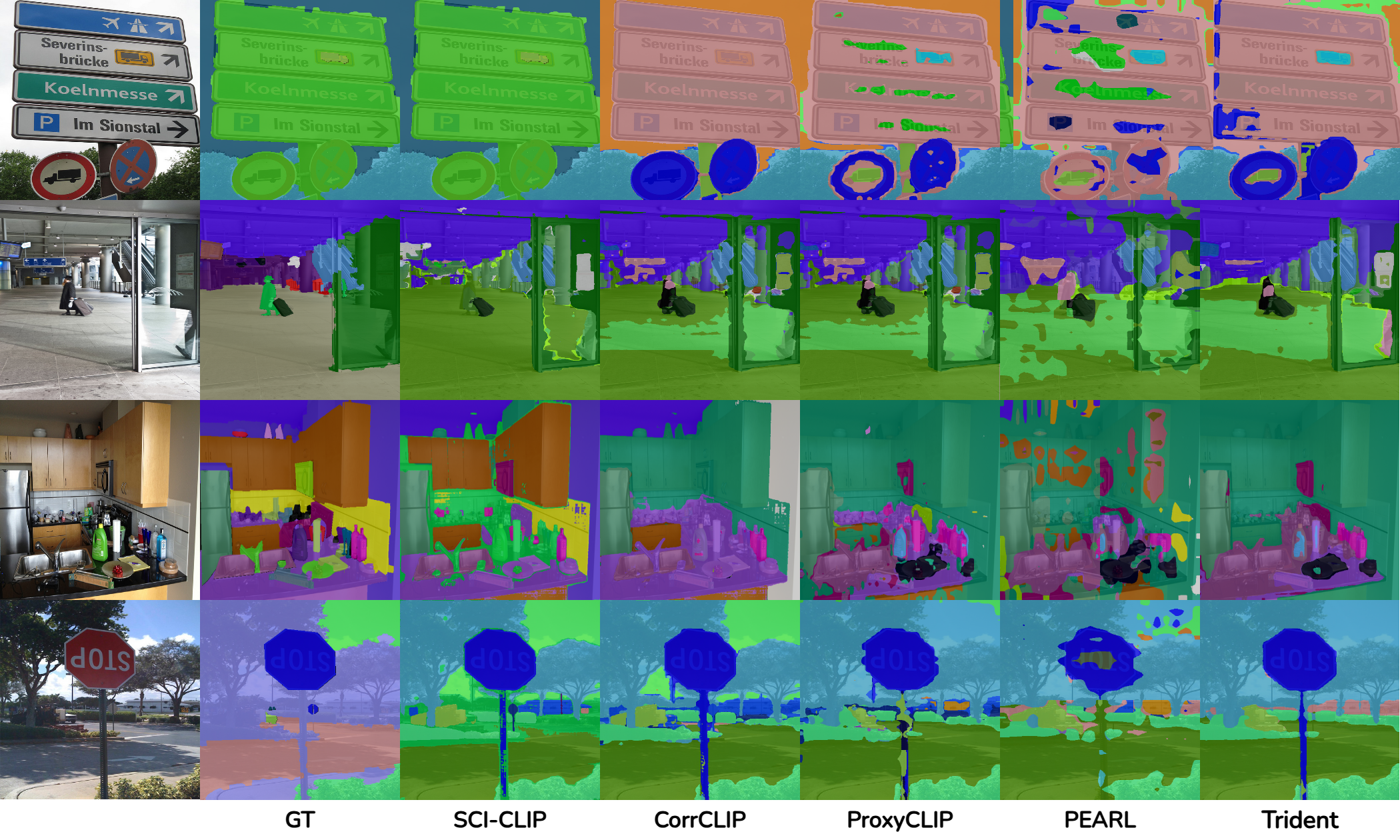}
    \caption{More qualitative comparison of our method, and the other four methods on COCO Stuff.}
    \label{fig:stuff}
\end{figure*}
\begin{figure*}[htbp]
    \centering
    \includegraphics[width=0.95\linewidth]{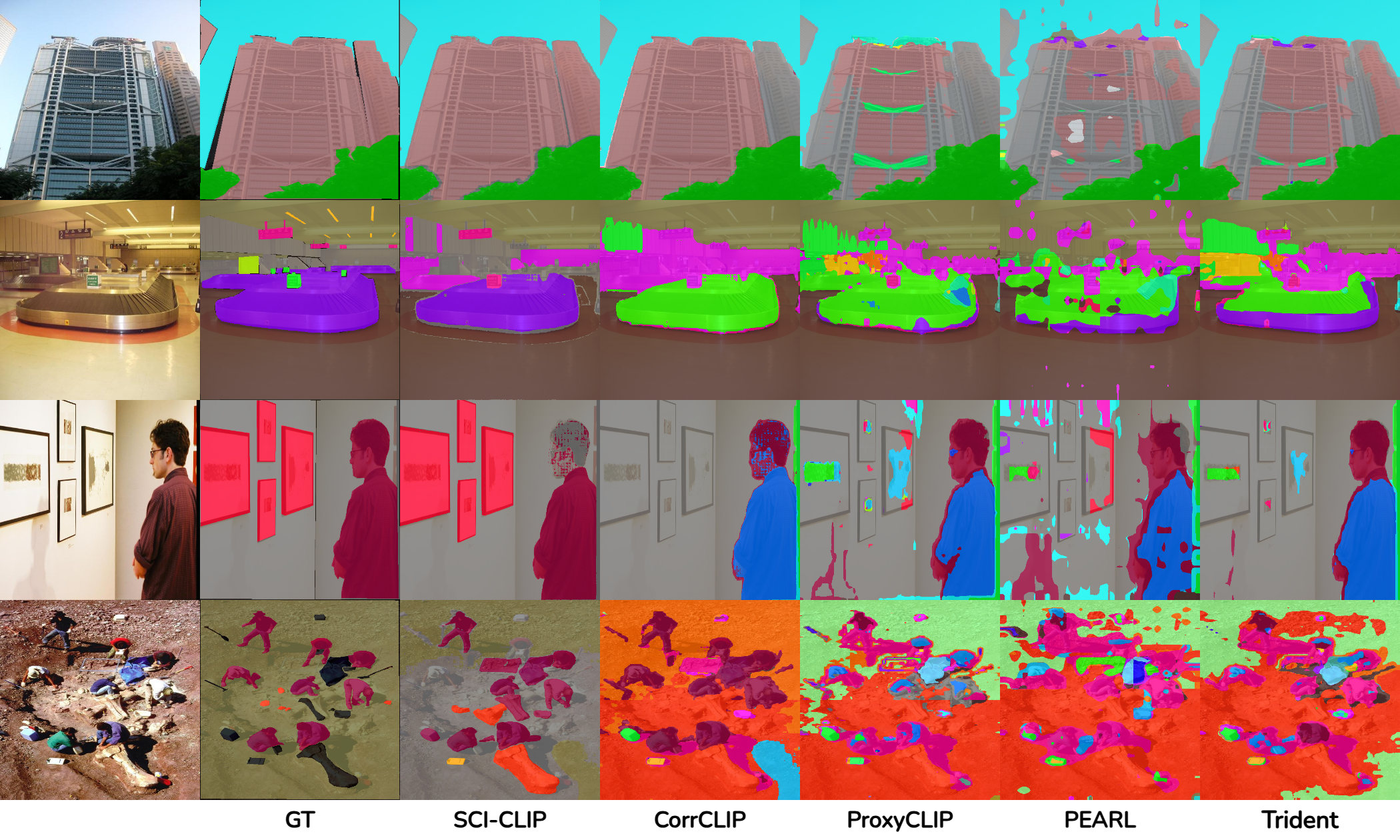}
    \caption{More qualitative comparison of our method, and the other four methods on ADE.}
    \label{fig:ade}
\end{figure*}

\label{sec:qualitatives}
We compare SCI-CLIP against prior segmentation approaches across diverse scenes and datasets, as illustrated in Figures~\ref{fig:city}, \ref{fig:stuff}, and \ref{fig:ade}. The results highlight consistent improvements in spatial coherence, object delineation, and robustness to scene complexity.

\paragraph{Structural Consistency vs. Fragmentation.}
Across driving scenes (Fig.~\ref{fig:city}), prior methods often produce fragmented predictions, particularly for thin structures such as poles and traffic signs, as well as distant objects. This behavior is commonly observed in approaches that rely primarily on local similarity or weak global priors. In contrast, SCI-CLIP maintains strong spatial continuity, generating coherent masks for roads, vehicles, and vertical structures. This improvement is enabled by its multi-stage refinement pipeline and the integration of global context with local evidence, which reduces patch-level inconsistencies.
\paragraph{Robustness to Dense and Cluttered Scenes.}
In cluttered indoor environments (Fig.~\ref{fig:stuff}), competing methods tend to either over-segment objects into multiple regions or under-segment by merging distinct categories, especially under high object density and appearance ambiguity. SCI-CLIP produces cleaner object boundaries and better separation between adjacent categories, while reducing noise in homogeneous regions such as walls and floors. This demonstrates stronger contextual reasoning beyond pixel-level similarity.

\paragraph{Fine-Grained Object Delineation.}
On challenging benchmarks such as ADE20K (Fig.~\ref{fig:ade}), prior methods struggle with thin structures, small objects, and scale variations. SCI-CLIP consistently preserves sharper edges, improves small-object recall, and better segments structurally complex regions. These results suggest that SCI-CLIP effectively captures multi-scale representations and avoids the over-smoothing artifacts observed in earlier methods.

\paragraph{Stability Under Appearance Variations.}
We observe that competing approaches are sensitive to illumination changes, texture variations, and viewpoint shifts, often leading to inconsistent predictions across similar scenes. In contrast, SCI-CLIP produces stable outputs across such variations, indicating improved feature alignment and a stronger reliance on semantic structure rather than raw appearance.

\paragraph{Failure Modes.}
Common failure patterns in prior methods include label bleeding between adjacent regions, noisy predictions in low-texture areas, and inconsistent object grouping. SCI-CLIP significantly mitigates these issues, although minor errors remain in highly ambiguous or heavily occluded regions.


\section{Leave-one-out ablation}
\label{sec:loo_ablation}
Table~\ref{tab:loo_ablation} confirms that the gains in Table \ref{tab:stagewise_ablation} are not merely artifacts of a particular construction order. In addition to mIoU, we also report the average memory footprint of each leave-one-out variant on NVIDIA 4$\times$L40s GPUs, so the it makes the accuracy--resource trade-off explicit at the module level. This additional column is informative because the SCI-CLIP modules do not all stress memory in the same way: scope reconstruction and Gaussian locality mainly reshape token interaction, whereas cross-window KV aggregation and reference-memory fusion introduce extra stored features, retrieval state, and multi-branch buffering during inference. Removing \emph{scope reconstruction} causes the largest drop in average mIoU, from $56.0$ to $53.1$, with especially severe degradation on City ($-5.6$), C60 ($-2.3$), and VOC21 ($-2.0$). This is consistent with the role of scope reconstruction as the structural backbone of SCI-CLIP: once region-constrained affinity is removed, the downstream modules operate on a noisier and less semantically coherent dense feature space.

\begin{table*}[htbp]
    \centering
    \small
    \setlength{\tabcolsep}{7pt}
    \caption{Leave-one-out ablation of SCI-CLIP. Starting from the full model, we remove one component at a time while keeping the remaining modules and dataset-specific hyperparameters fixed. }
    \begin{tabular}{lcccccccc}
        \toprule
        Variant & C60 & City & Object & Stuff & V20 & V21 & Avg. &Memory \\
        \midrule
        Full SCI-CLIP & \underline{46.1} & \textbf{44.7} & \textbf{45.2} & \underline{33.8} & \underline{90.3} & \underline{75.9} & \textbf{56.0} &4585.2\\
        w/o Scope & 43.8 & 39.1 & 41.2 & 31.6 & 89.2 & 73.9 & 53.1&4566.5 \\
        w/o Mask Merging & 46.1 & 44.7 & 43.4 & \textbf{33.9} & \textbf{90.4} & 75.8 & \underline{55.7} &4585.3\\
        w/o Gaussian & 45.9 & 44.1 & \underline{43.5} & 33.7 & 89.9 & \textbf{76.1} & 55.6 &4569.1\\
        w/o Cross-Window KV & \textbf{46.2} & 42.5 & 43.4 & 33.9 & 90.1 & 75.4 & 55.2&3332.7 \\
        w/o Ref. Memory & 42.1 & \underline{44.2} & 43.4 & 30.9 & 88.9 & 72.8 & 53.7&3972.8 \\
        \bottomrule
    \end{tabular}
    \label{tab:loo_ablation}
\end{table*}

The leave-one-out study also clarifies the roles of the later modules. \emph{Mask merging} has only a marginal effect on the final average, indicating that its value is mostly as a stabilizer rather than a dominant source of gains; correspondingly, its memory effect should be small because it changes the region partition more than the volume of stored activations. \emph{Gaussian prior} is similarly modest in aggregate, which matches the stagewise observation that locality bias is helpful but not sufficient on its own; it adds structured bias to affinity rather than a new feature bank, so its memory cost is expected to be limited. In contrast, removing \emph{cross-window KV aggregation} reduces the average to $55.2$ and causes a pronounced drop on City, reinforcing the claim that crop-boundary consistency requires an explicit long-range correction path beyond local scope reconstruction. This module is also the one for which a memory increase is most intuitive, since it must maintain cross-window feature state and retrieval buffers that do not exist in a purely local variant.

Finally, removing \emph{reference memory} lowers the average from $56.0$ to $53.7$ ($-2.1$), with the largest losses on C60, V21, and Stuff. This pattern is highly informative: the reference memory is not uniformly beneficial across all categories, but it is particularly valuable on benchmarks where dense open-vocabulary predictions remain ambiguous and exemplar-based correction can disambiguate semantically similar regions. Its memory behavior is also intuitive: unlike the purely dense pathway, the memory-enabled variant must store and query an external bank of segment embeddings and label affinities, so improved accuracy comes with additional retrieval-time state. Together with the appended memory measurement, this gives a more complete view of the design space: the leave-one-out ablation not only identifies which modules are most important for accuracy, but also clarifies why some modules are more resource-intensive than others. Tables ~\ref{tab:loo_ablation} and \ref{tab:stagewise_ablation} support a coherent picture of SCI-CLIP. Scope reconstruction is the primary structural ingredient, cross-window KV aggregation is the key long-range contextual mechanism, and reference memory provides the largest late-stage semantic correction, while mask merging and Gaussian locality act as lower-level refinements that improve stability around that core pipeline.

\end{document}